\documentclass[runningheads]{llncs}

\usepackage{eccv}

\usepackage{eccvabbrv}

\usepackage{graphicx}
\usepackage{booktabs}

\usepackage[accsupp]{axessibility}  

\usepackage{algorithm}
\usepackage{algorithmic}
\usepackage{multirow}
\usepackage{colortbl}
\usepackage[normalem]{ulem}
\useunder{\uline}{\ul}{}
\usepackage{marvosym}

\newcommand{\zjk}[1]{{#1}}

\usepackage{hyperref}

\usepackage{orcidlink}

\begin{document}

\title{CoSIGN: Few-Step Guidance of ConSIstency Model to Solve General INverse Problems} 

\titlerunning{CoSIGN}

\author{Jiankun Zhao\inst{1}\thanks{Equal contribution.} \and
Bowen Song\inst{1}$^{\star}$\orcidlink{0009-0005-1285-3048} \and
Liyue Shen\inst{1}\Letter\orcidlink{0000-0001-5942-3196}}

\authorrunning{J.~Zhao et al.}

\institute{University of Michigan \\
\email{jkzhao.nku@gmail.com\quad\{bowenbw, liyues\}@umich.edu}}

\maketitle

\begin{abstract}
Diffusion models have been demonstrated as strong priors for solving general inverse problems. Most existing Diffusion model-based Inverse Problem Solvers (DIS) employ a plug-and-play approach to guide the sampling trajectory with either projections or gradients. Though effective, these methods generally necessitate hundreds of sampling steps, posing a dilemma between inference time and reconstruction quality. In this work, we try to push the boundary of inference steps to 1-2 NFEs while still maintaining high reconstruction quality. To achieve this, we propose to leverage a pretrained distillation of diffusion model, namely consistency model, as the data prior. The key to achieving few-step guidance is to enforce two types of constraints during the sampling process of the consistency model: \textit{soft measurement constraint} with ControlNet and \textit{hard measurement constraint} via optimization. 
Supporting both single-step reconstruction and multistep refinement, the proposed framework further provides a way to trade image quality with additional computational cost. 
Within comparable NFEs, our method achieves new state-of-the-art in diffusion-based inverse problem solving, showcasing the significant potential of employing prior-based inverse problem solvers for real-world applications.
Code is available at: \url{https://github.com/BioMed-AI-Lab-U-Michgan/cosign}.


\end{abstract}
\vspace{-12pt}
\section{Introduction}\label{sec:intro}

Inverse problems cover a wide spectrum of fundamental image restoration tasks, such as inpainting, super-resolution and deblurring~\cite{mcg,dps,pigdm,lgd-mc,ddrm,resample,cddb}. These problems aim at recovering the original signals $\boldsymbol{x}$ given downgraded measurements $\boldsymbol{y}$. A forward operator $\mathcal{A}$, either linear or nonlinear, determines the process transforming original signal to downgraded measurements. Due to the sparse sampling nature of this process,  $\mathcal{A}$ is analytically irreversible, causing the problem to be ill-posed~\cite{score-sde,ddnm}.

\begin{figure}[t]
  \centering
   \includegraphics[width=.75\linewidth]{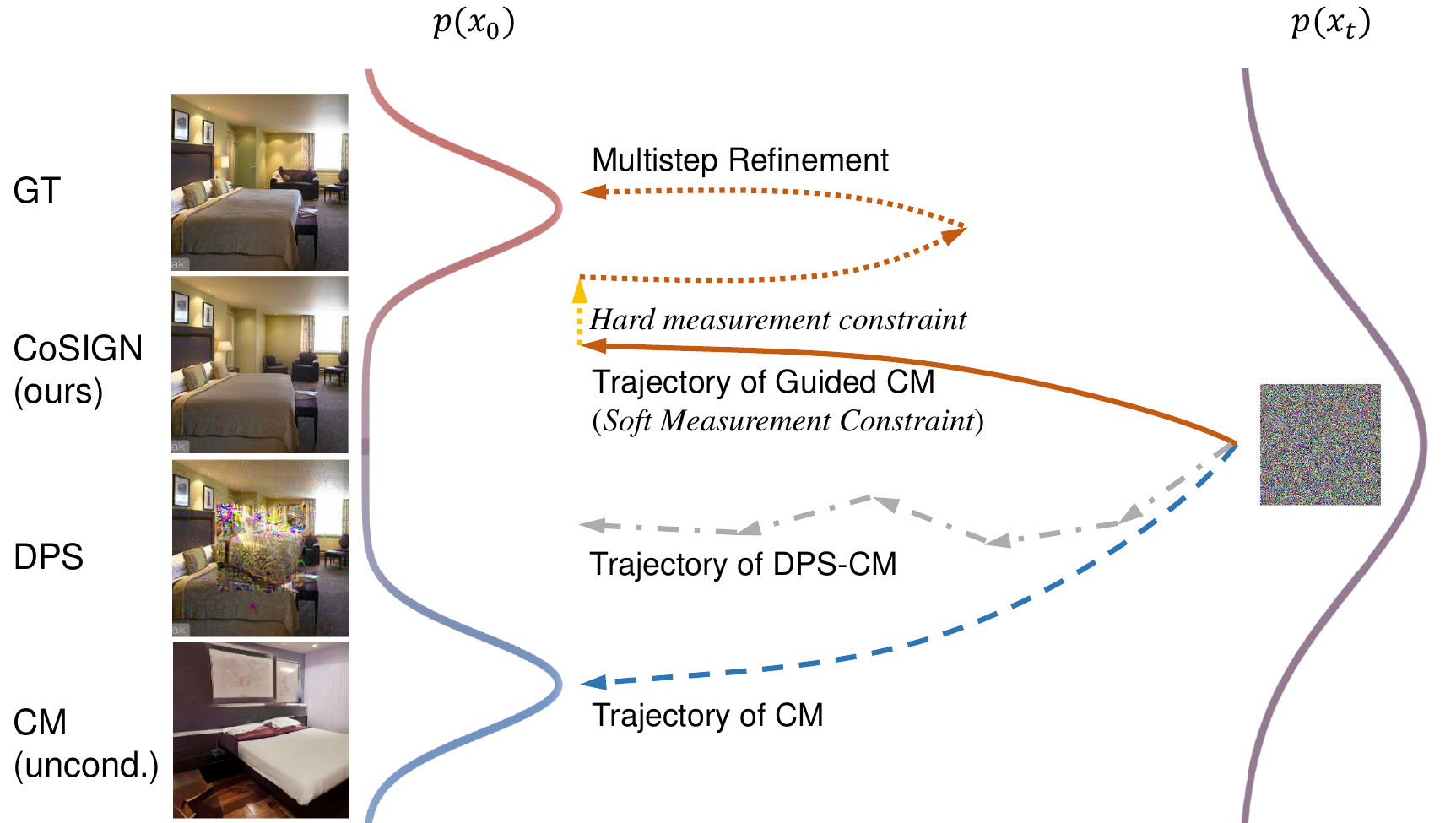}
    \vspace{-8pt}
   \caption{Illustration of the sampling trajectory and reconstruction results. In low NFE region, typical DIS methods like DPS (grey arrow) fail to guide original trajectory of CM (blue arrow) towards high-fidelity results. Instead, our method (orange arrow) can guide the trajectory in a single step with ControlNet, and further refine the single-step result with hard measurement constraint and multistep sampling.}
   \vspace{-20pt}
   \label{fig:intro}
\end{figure}


Traditionally, one can solve inverse problems with mathematical regularization~\cite{sparse,low-rank}, or train an end-to-end neuron network to map the measurement to its corresponding original signal~\cite{swinir,deepfill}. However, due to lack of creativity, these methods generally suffer from over-smoothness and blurry artifacts. Emergence of powerful deep generative models~\cite{vae,gan,score_orig} provided a new approach to remedy information lost in the forward process. This concept was first applied in an unsupervised manner, where an unconditional generative model is utilized as a prior. By incorporating the prior with various measurement constraints, previous works on this line~\cite{gan_cs,pulse,gan_inv} explored methods to sample from posterior distribution $p(\boldsymbol{x}|\boldsymbol{y})$ given pretrained unconditional generative models. Recent works~\cite{score-sde,ddrm,mcg} have found diffusion models particularly suitable as generative prior in this approach. Not only can diffusion models generate high-fidelity samples without adversarial training, but their iterative sampling process is also naturally compatible with plug-and-play measurement constraints. A batch of these methods, named Diffusion-based Inverse problem Solvers (DIS)~\cite{dps,ddnm,pigdm,lgd-mc}, were developed to "hijack" the sampling trajectory towards measurement-consistent samples. However, reliance on hundreds of iterative sampling steps also limits their application in real-time or high-dimensional scenarios, such as video processing and 3D imaging~\cite{vidm,dreamfusion,prolificdreamer}. Inspired by the success of unsupervised DISs, another line of works~\cite{palette,i2sb,cddb} try to utilize diffusion models to solve inverse problems in a supervised manner. These works directly model the posterior distribution $p(\boldsymbol{x}|\boldsymbol{y})$ by taking the measurements as inputs during model training. Rather than utilizing an off-the-shelf generative prior, these methods train a generative model from scratch for each task, which \zjk{limits their performance on out-of-domain inverse tasks.} 
Besides, these methods also rely on iterative sampling to generate high-fidelity samples.

To address these challenges, continuous efforts have been made to reduce sampling steps of diffusion models. One promising approach is by distilling a pretrained diffusion model into an implicit model that directly maps noise to samples~\cite{progdist,on_distill,cm,lcm}. Among them, Consistency Model (CM)~\cite{cm} was recently proposed as an efficient distillation method to enable single-step sampling of diffusion-based generative models. Despite the powerful few-step generation ability of CM, it is not a trivial task to directly leverage CM as a prior for solving inverse problems. In~\cref{fig:intro}, we illustrate why existing DISs, \zjk{including the inverse problem solving algorithm proposed in \cite{cm}}, fail to guide the unconditional sampling trajectory of CM towards a measurement-consistent sample under the few-step setting. \zjk{CM prior differs from diffusion prior in that it predicts $x_0$ rather than their expectation $\hat{x_0}$, whereas most DISs rely on $\hat{x_0}$ to approximate the likelihood score $\nabla_{\boldsymbol{x}_t} \log p_t\left(\boldsymbol{y}|\boldsymbol{x}_t\right)$. Inaccuracy in this approximation leads the sample away from the authentic data distribution~\cite{lgd-mc, pigdm}, leaving it an open problem to incorporate the CM priors into few-step inverse problem solvers.}

In this work, we focus on tackling the challenge of long sampling process in previous methods. Motivated by CM, we propose CoSIGN, a few-step guidance method of ConSIstency model to solve General INverse
problems. CoSIGN utilize the strong data prior from CM with improved sampling efficiency to generate results in only a single or few steps. To be specific, we first propose to guide the sampling process of consistency model with a \textit{soft measurement constraint}: training an additional encoder, namely ControlNet~\cite{controlnet}, over the Consistency Model backbone. The ControlNet takes the downgraded measurement or its pseudo-inverse as the input, and controls the CM output to be consistent with the measurement. With the pretrained consistency model as a frozen backbone, we only need to train the ControlNet to guide the sampling process of CM in a single step (see ~\cref{fig:intro}).
Moreover, to further improve fidelity of the generated reconstructions, we propose to plug in a \textit{hard measurement constraint} module to explicitly guarantee measurement consistency and reduce distortion. 
The proposed framework is capable of solving linear, nonlinear, noisy and blind inverse problems, as long as \zjk{paired data can be constructed to train the ControlNet.}

Our main contributions can be concluded as follows:
\begin{itemize}
\item We propose a few-step inverse problem solver with improved sampling efficiency by leveraging the pretrained consistency model as a data prior.
\item We propose to guide the conditional sampling process of consistency models with both soft measurement constraint and hard measurement constraint, which enables generating high-fidelity, measurement-consistent reconstructions within 1-2 NFEs;
\item Experiments demonstrate superior few-step reconstruction ability of our method on four tasks of linear and non-linear inverse problems, where our method achieves state-of-the-art within low NFE regime, while competitive with those methods generated using about 1000 NFEs.
\end{itemize}

\section{Related Work}\label{sec:relative_work}
\noindent\textbf{Generative Inverse Problem Solvers.} 
Deep learning based approaches for solving inverse problems can be generally categorized into two types: supervised methods and unsupervised methods~\cite{deep_inverse, ir_review}. Generative models were first introduced into unsupervised approaches as a data prior. Early works~\cite{gan_cs, gan_inv, pulse} explore the latent space of Generative Adversarial Networks (GAN)~\cite{gan}, guided by gradients or projections towards the measurement. 
With the emergence of generative diffusion models, recent works~\cite{score-sde, mcg} integrated such guidance with the iterative sampling process of diffusion models~\cite{score, ddpm, ddim, adm} for solving inverse problems. These methods, dubbed Diffusion-base Inverse problem Solvers (DIS), bifurcated into two main streams based on ways to enforce measurement consistency. Soft approaches~\cite{dps, lgd-mc, pigdm} attempted to keep samples on generative manifolds by taking gradients through the neural network, while hard approaches~\cite{ddrm, ddnm, dds} project the measurement back to the signal space and replace "range space"~\cite{ddnm} of current sample with the projection. \zjk{Some recent studies explored more accurate posterior approximation with normalizing flow~\cite{flow} or partical filtering~\cite{fps}.} But most DIS methods, no matter hard or soft, requires a lot of sampling steps to obtain measurement-consistent samples.


Advancements of unsupervised DIS methods have inspired the use of probabilistic generative models under supervised settings. Instead of fitting $p(\boldsymbol{x})$, Palette~\cite{sr_dif, palette} trained an image-to-image conditional diffusion model that directly fits $p(\boldsymbol{x}|\boldsymbol{y})$. I$^2$SB~\cite{i2sb} and CDDB~\cite{cddb} algorithms narrowed the gap between $\boldsymbol{x}_T$ and $\boldsymbol{x}_0$ by building a Schr{\"o}dinger Bridge, improving reconstruction quality in extremely low NFE (number of function evaluations) regime of less than 10. Compared with end-to-end supervised approaches mapping $\boldsymbol{y}$ to $\boldsymbol{x}$~\cite{srgan,deepfill,swinir}, these methods significantly improved image quality due to diffusion model priors. 
\zjk{However, since the entire model is trained for a particular task from scratch, these methods usually generalize poorly on out-of-domain tasks.}

\noindent\textbf{Guiding Diffusion Models.} Existing works have explored methods to guide diffusion models with class labels~\cite{adm}, texts~\cite{ldm,imagen,glide} and images~\cite{ldm,controlnet}. Classifier guidance~\cite{adm} and Classfier-Free Guidance (CFG)~\cite{cfg} pioneered class-conditioned generation. Benefiting from CFG, large text-to-image diffusion models like Latent Diffusion~\cite{ldm}, Imagen~\cite{imagen} and Glide~\cite{glide} enjoyed great success by leveraging pre-trained text and image encoder to inject guidance into sampling process of diffusion models. ControlNet~\cite{controlnet} provided a fine-tuning method to adapt these pre-trained models to specific image-to-image translation tasks. In tasks like sketch-to-image and depth-to-image, ControlNet showed superior ability in enhancing semantic and structural similarity. However, its ability of enforcing measurement consistency in general inverse problems still remains under-explored.

\noindent\textbf{Accelerating Diffusion Models.} Slow sampling speed has been limiting the real-world applications of diffusion models in generation of 3D scenes~\cite{dreamfusion, prolificdreamer}, videos~\cite{vidm} and speeches~\cite{comospeech}. Recent works proposing to switch from SDE~\cite{ddpm} to ODE~\cite{ddnm} and adopt higher-order ODE solvers~\cite{edm, dpm, dpm++, dpmv3} managed to accelerate sampling process to 5-10 NFEs. Unfortunately, these ODE solvers cannot be directly applied to the likelihood score $\nabla_{\boldsymbol{x}_t} \log p_t\left(\boldsymbol{y} | \boldsymbol{x}_t\right)$ in DIS. Another line of works~\cite{progdist, on_distill, cm, lcm} explored methods to distill the sampling trajectory of pre-trained diffusion models. Although these methods efficiently accelerate unconditional image generation, how can these distillation models help accelerate inverse problem solving has not been thoroughly studied yet.
\section{Background}\label{sec:bg}

\noindent\textbf{Inverse Problem Solving} aims at reconstructing an unknown signal $\boldsymbol{x}\in\mathbb{R}^n$ based on the measurements $\boldsymbol{y}\in\mathbb{R}^m$ . Formally, $\boldsymbol{y}$ derives from a forward process determined by ~\cref{eq:inverse},
\begin{equation}
  \boldsymbol{y} = \mathcal{A}(\boldsymbol{x})+\boldsymbol\epsilon
  \label{eq:inverse}
\end{equation}
where $\mathcal{A}$ can be either a linear operator like Radon transformation in sparse-view CT reconstruction, or a nonlinear operator like JPEG restoration encoder. $\boldsymbol\epsilon$ denotes random noise along with the measurement acquisition. Inverse problem becomes ill-posed when it comes to a case where $m<n$. In other words, $\mathcal{A}^{-1}$ does not exist, thus there are multiple $\boldsymbol{x}$ satisfying ~\cref{eq:inverse}. In this case, some format of data prior is required to recover the original signal $\boldsymbol{x}$.
Solving inverse problem is to find an optimal $\boldsymbol{x}$ that is both consistent with the measurement $\boldsymbol{y}$ and the prior knowledge of $p(\boldsymbol{x})$. The solution can be usually formulated as:
\begin{equation}
\hat{\boldsymbol{x}}=\underset{\boldsymbol{z} \in \mathbb{R}^n}{\arg \min }\left\{\|\mathcal{A}(\boldsymbol{z})-\boldsymbol{y}\|^2+\rho \mathcal{R}(\boldsymbol{z})\right\} \quad 
  \label{eq:inverse_solving}
\end{equation}
where the first term $\|\mathcal{A}(\boldsymbol{z})-\boldsymbol{y}\|^2$ optimizes results towards measurement consistency and the second term $\rho \mathcal{R}(\boldsymbol{z})$ regularizes the result with the knowledge of $p(\boldsymbol{x})$.  Generative models can also be utilized as the data prior in the regularization term. For example, diffusion models can be trained to capture the prior data distribution $p(x)$ via score matching~\cite{score_orig}. In this case, splitting prior score from posterior with Bayes' Rule is equivalent to solving~\cref{eq:inverse_solving} with gradient descent:\footnote{$t\in\left[0, T\right]$ denote the index of a particular noise level in diffusion model. Please refer to ~\cite{score_orig} for details about diffusion models.}
\begin{equation}
\nabla_{\boldsymbol{x}_t} \log p_t\left(\boldsymbol{x}_t|\boldsymbol{y}\right)=\nabla_{\boldsymbol{x}_t} \log p_t\left(\boldsymbol{y}|\boldsymbol{x}_t\right)+\nabla_{\boldsymbol{x}_t} \log p_t\left(\boldsymbol{x}_t\right)
  \label{eq:inverse_prob}
\end{equation}
But this equivalency only holds when assuming\cite{dps}:
\begin{equation}
\nabla_{\boldsymbol{x}_t} \log p_t\left(\boldsymbol{y}|\boldsymbol{x}_t\right)\simeq-\frac{1}{\lambda}\nabla_{\boldsymbol{x}_t}\|\mathcal{A}(\hat{\boldsymbol{x}}_0(\boldsymbol{x}_t))-\boldsymbol{y}\|^2
\label{eq:approx}
\end{equation}
where \zjk{$\lambda$ is a hyperparameter controlling step size, and} $\hat{\boldsymbol{x}}_0(\boldsymbol{x}_t)$ denotes an estimation of clean sample $\boldsymbol{x}_0$ based on intermediate $\boldsymbol{x}_t$ with Tweedie's formula. Such approximation brings significant error when discretizing the diffusion process into only a few steps. 
This motivates us to turn a pre-trained diffusion model fitting $p(\boldsymbol{x})$ into one that directly fits $p(\boldsymbol{x}|\boldsymbol{y})$, instead of approximating likelihood hundreds of times during sampling process.


\noindent\textbf{Consistency Model} (CM) is a new family of generative models distilled from diffusion models that can generate high-quality image samples in a single step. Furthermore, these samples can be refined through multistep sampling of CM. Rather than estimating $\mathbb{E}(\boldsymbol{x}_0|\boldsymbol{x}_t)$ as diffusion models, CM aims at learning a direct mapping from any point on an ODE trajectory to its origin, i.e., $f_\theta: (\boldsymbol{x}_t, t) \mapsto \boldsymbol{x}_\epsilon$\footnote{$\epsilon$ is a sufficient small positive number to stabilize training.}.

The key to successfully building $f$ is enforcing two constraints: self-consistency and boundary constraint. Self-consistency states that for all points on a particular trajectory, $f$ must output the same origin point $\boldsymbol{x}_0$ (see ~\cref{eq:self_consistency}).
\begin{equation}
f_\theta(\boldsymbol{x}_t, t)=f_\theta(\boldsymbol{x}_{t^\prime}, t^\prime) \quad \forall t, t^\prime\in[\epsilon, T]
\label{eq:self_consistency}
\end{equation}
Meanwhile, boundary constraint ensures that this original point is the authentic data itself (see ~\cref{eq:boundary_constraint}).
\begin{equation}
f_\theta(\boldsymbol{x}_\epsilon, \epsilon)=\boldsymbol{x}_\epsilon
\label{eq:boundary_constraint}
\end{equation}
Inspired by \cite{edm}, boundary constraint is enforced by introducing skip connection:
\begin{equation}
f_\theta(\boldsymbol{x}, t)=c_{skip}(t)\boldsymbol{x}+c_{out}(t)F_\theta(\boldsymbol{x}, t)
\label{eq:skip}
\end{equation}
where $c_{skip}(t)$ and $c_{out}(t)$ are specially designed functions satisfying $c_{skip}(\epsilon)=1$ and $c_{out}(\epsilon)=0$, while $F_\theta(\boldsymbol{x}, t)$ is a U-Net parameterized by $\theta$.

A CM can be obtained by either distilling a pre-trained diffusion model (dubbed consistency distillation, CD) or training from scratch (dubbed consistency training, CT). In both ways, loss functions are designed to guarantee self-consistency. Specifically, these two losses have the same form of\footnote{$f_\theta$ denotes the "online network" while $f_{\theta^{-}}$ denotes the "target network". Please refer to \cite{cm} for details about exponential moving average (EMA).}:
\begin{equation}
\mathcal{L}_{cm}=d(f_\theta(\boldsymbol{x}_{t_{n+1}}, t_{n+1}), f_{\theta^{-}}(\widetilde{\boldsymbol{x}}_{t_n}, t_n))
\label{eq:cm_loss}
\end{equation}
where $d(\cdot, \cdot)$ is a distance function like L1, L2 or LPIPS\cite{lpips}. In CD, $\widetilde{\boldsymbol{x}}_{t_n}$ is predicted by the pre-trained diffusion model based on $x_{t_{n+1}}$. Whereas in CT, $\widetilde{\boldsymbol{x}}_{t_n}$ represents $\boldsymbol{x}_{t_n}$ perturbed by the same noise $\epsilon\sim\mathcal{N}(0, I)$ as $\boldsymbol{x}_{t_{n+1}}$. We choose CD in this work since distilled trajectories are closer to those in diffusion models, and thus can generate images with better quality.
\section{Method}\label{sec:method}

\begin{figure*}[t]
  \centering
   \includegraphics[width=\linewidth]{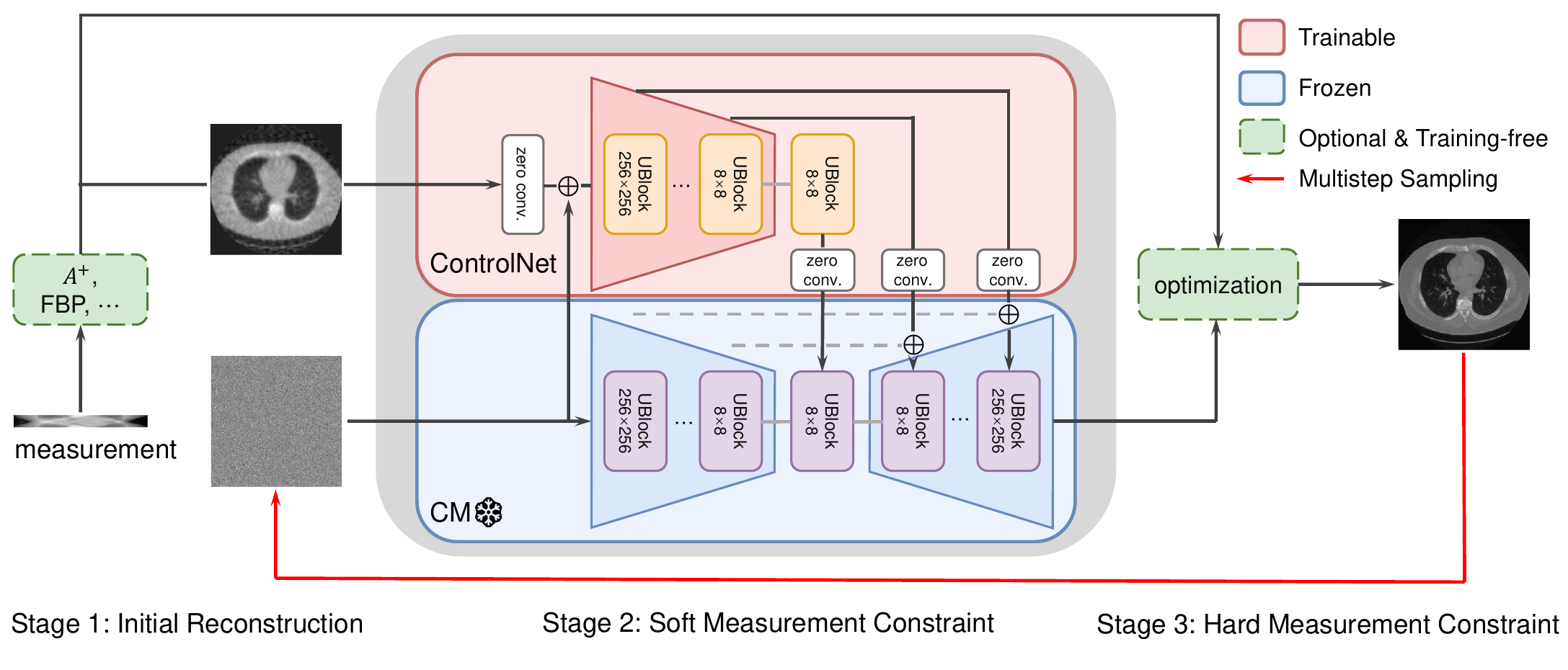}
   \vspace{-16pt}
   \caption{Overview of our proposed CoSIGN method. In Stage 1, measurement is projected onto the signal space through initial reconstruction. In Stage 2, we input the initial reconstruction into the ControlNet as a condition, and guide the pretrained CM with soft measurement constraint. In Stage 3, we further guarantee measurement consistency with hard measurement constraint. Both the first and the third stage are optional and training-free. 
   In the multistep sampling scheme, 
   the single-step reconstruction result can be sent back to the second stage for further refinement by adding a lower level of noise and denoising with CM for a second time.}
   \vspace{-18pt}
   \label{fig:method}
\end{figure*}

In this section, we will discuss how we enable few-step inverse problem solving with pre-trained consistency models as data priors. 
Firstly, different from previous works which aims at improving reconstruction quality in low NFE region~\cite{ddrm,i2sb}, we are motivated to develop a method to enable high-fidelity single-step reconstruction. 
To be specific, as shown in~\cref{fig:method}, we propose to control the sampling process of consistency model with a \textit{soft measurement constraint} by building a ControlNet~\cite{controlnet} over the consistency model backbone (\cref{sec:controlnet}). 
To feed the measurement into the ControlNet as a condition, we reformulate the inverse problems by adding an initial reconstruction stage to transform the measurements in different inverse problems into the signal space.
Secondly, while most image-to-image translation tasks mainly seek semantic similarity, inverse problems require strict measurement consistency. 
Thus, we further introduce \textit{hard measurement constraint} in the multistep sampling scheme to explicitly guarantee consistency with the measurements in a more robust way (\cref{sec:multistep}).

\subsection{Soft Measurement Constraint via ControlNet}
\label{sec:controlnet}
In~\cref{sec:bg}, we introduced two essential constraints when training consistency models for unconditional image generation: self-consistency and boundary constraint. Under this setting, it is expected that each $\boldsymbol{x}_t$ belong to only one trajectory, and this trajectory ends at a deterministic point~\cite{cm}. However, when solving inverse problems, which trajectory an $\boldsymbol{x}_t$ belongs to depends on not only $\boldsymbol{x}_t$ itself but also the measurements $\boldsymbol{y}$. In other words, we can expand~\cref{eq:self_consistency} to:
\vspace{-5pt}
\begin{equation}
f_\theta(\boldsymbol{x}_t, \boldsymbol{y}, t)=f_\theta(\boldsymbol{x}_{t^\prime},  \boldsymbol{y}, t^\prime) \quad \forall t, t^\prime\in[\epsilon, T]
\label{eq:measurement_consistency}
\vspace{-5pt}
\end{equation}
One straightforward way to satisfy ~\cref{eq:measurement_consistency} is to train or distill a conditional consistency model with~\cref{eq:consistency_loss}, where $\widetilde{\boldsymbol{x}}_{t_n}$ is derived by adding the same noise to $\boldsymbol{x}_{0}$ under the training setting, or predicted by the teacher diffusion model under the distilling setting.
\vspace{-5pt}
\begin{equation}
\mathcal{L}_{con}=d(f_\theta(\boldsymbol{x}_{t_{n+1}}, \boldsymbol{y}, t_{n+1}), f_{\theta^{-}}(\widetilde{\boldsymbol{x}}_{t_n}, \boldsymbol{y}, t_n))
\label{eq:consistency_loss}
\vspace{-5pt}
\end{equation}
But we empirically find optimizing $\mathcal{L}_{con}$ leads to slow convergence. Hence, we are motivated to circumvent this problem by exploring other ways to ensure $\mathcal{A}(f(\boldsymbol{x}_t, \boldsymbol{y}, t))=\boldsymbol{y}$. Specifically, we introduce \textit{soft measurement constraint} in addition to self-consistency. One can simply enforce soft measurement constraint with the following loss:
\vspace{-5pt}
\begin{equation}
\mathcal{L}_{recon}=d(f_\theta(\boldsymbol{x}_t, \boldsymbol{y}, t), \boldsymbol{x}_0)
\label{eq:loss}
\vspace{-5pt}
\end{equation}
where $x_0$ is the ground-truth. Note that $f_\theta(\boldsymbol{x}_t, \boldsymbol{y}, t)=\boldsymbol{x}_0$ is a stronger condition than ~\cref{eq:measurement_consistency}. Training a U-Net from scratch with $\mathcal{L}_{recon}$ will encourage $f_\theta(\boldsymbol{x}_t, \boldsymbol{y}, t)$ to approximate $\mathbb{E}[\boldsymbol{x}_0|\boldsymbol{x}_t, \boldsymbol{y}]$, resulting in a diffusion model rather than a consistency model.
 
To strike a balance between soft measurement constraint and self-consistency, we follow~\cite{controlnet} to freeze the unconditional CM backbone while train an additional encoder (a.k.a. ControlNet) over it to guide the sampling process of CM and enable conditional generation. 
Since we explicitly keep the CM backbone frozen, the single-step generation ability can be inherited from pretrained CM even if we train the ControlNet with $\mathcal{L}_{recon}$. 
The structure of the whole model is depicted in Stage 2 of ~\cref{fig:method}. The encoder and the middle block of the ControlNet share a same architecture as their counterparts in CM backbone. Their parameters are also initialized with those in the pretrained CM backbone. In this way, ControlNet inherit the outstanding perceptual ability of pretrained CM. Additionally, zero convolutions are added to the inlet and outlet of ControlNet to avoid disturbance from random noise in the initial training stage. By training this elaborately initialized network with $\mathcal{L}_{recon}$ defined in ~\cref{eq:loss}, our model progressively learns to be consistent with the condition, maintaining the single-step generation ability at the same time.

It is worth noticing that ControlNet is originally designed to deal with image-to-image translation tasks. Whereas in many inverse problems, the measurements $\boldsymbol{y}$ has a different size or even lies in a different space from the signals $\boldsymbol{x}$. For instance, in sparse-view CT reconstruction, the measurement is a sinogram rather than an image (see Stage 1 of~\cref{fig:method}). Thus, we introduce an initial reconstruction stage before the guided consistency model to adapt inverse problems as an input of ControlNet. Specifically, we categorize inverse problems into three types, and adopt different initial reconstruction methods for each type. First, all linear inverse problems have a pseudo-inverse operator $\boldsymbol{A}^\dag$ satisfying $\boldsymbol{A}\boldsymbol{A}^\dag\boldsymbol{A}\equiv\boldsymbol{A}~$\cite{ddnm}. 
In tasks like inpainting and super-resolution, $\boldsymbol{A}^\dag$ can be easily formed as $\boldsymbol{A}^\dag=\boldsymbol{A}$ and $\boldsymbol{A}^\dag\in\mathbb{R}^{s^2\times 1}=\left[1, \cdots, 1 \right]^T$ respectively, where $s$ denotes the super-resolution scale. However, in other linear tasks like sparse-view CT reconstruction, $\boldsymbol{A}^\dag$ can only be constructed by SVD and complex Fourier transform~\cite{ddnm, ddrm}, or estimated by conjugated gradient~\cite{dds, resample}. Instead of using these time-consuming methods to derive $\boldsymbol{A}^\dag$, we can use more efficient reconstruction methods like filtered back-projection (FBP), which is a commonly-used standard way for CT reconstruction. 
Lastly, for nonlinear problems and problems with unknown forward operator, we merely input the resized measurement as condition and let ControlNet learn a transition from measurement to signal.

\subsection{Hard Measurement Constraint and Multistep Sampling}
\label{sec:multistep}

While most image-to-image translation tasks mainly seek semantic similarity, inverse problems require strict measurement consistency. 
Thus, we further introduce \textit{hard measurement constraint} in the multistep sampling scheme to explicitly guarantee consistency with the measurements in a more robust way. \zjk{Hard measurement constraint is an optimization step that guides the sample to be consistent with the measurement. }
Theoretically, most \zjk{optimization methods in} existing DISs, either soft or hard, can be applied here as a plug-and-play module. But we empirically found hard approaches more effective under the few-step setting. The goal is to find the projection of the prior "clean" sample $\boldsymbol{x}_0$ to the manifold that is measurement-consistent.  Let $\epsilon$ be the tolerance threshold for the noise in the measurement $\boldsymbol{y}$, the optimization objective is given by 
\begin{equation}
\widehat{\boldsymbol{x}}_{0}=\underset{\boldsymbol{z} \in \mathbb{R}^n}{\arg \min}\left\{\|\boldsymbol{z}-\boldsymbol{x}_0\|_2^2\right\} \quad \text{s.t.} \quad \|\mathcal{A}(\boldsymbol{z}) - \boldsymbol{y}\|_2^2 \leq \epsilon^2.
  \label{eq:obj_piccs}
\end{equation}
and the Lagrangian form of~\cref{eq:obj_piccs} is given by $\|\boldsymbol{z}-\boldsymbol{x}_0\|_2^2 + \varphi \|\mathcal{A}(\boldsymbol{z}) - \boldsymbol{y}\|_2^2 $. 
\zjk{For linear inverse problems}, the previous optimization objective can be solved with a close-form solution by computing the pseudo-inverse \cite{ddnm}. We adopt the relaxed projection form in DDNM~\cite{ddnm} \zjk{as an example in our experiment}, which updates the "clean" sample $\boldsymbol{x}_{0}$ with~\cref{eq:ddnm}.
\begin{equation}
\widehat{\boldsymbol{x}}_{0}=\boldsymbol{x}_{0} + \kappa(\boldsymbol{A}^\dag \boldsymbol{y} - \boldsymbol{A}^\dag \boldsymbol{A}\boldsymbol{x}_{0})
\label{eq:ddnm}
\footnote{\zjk{$\kappa$ is a hyperparameter introduced in DDNM+ to control noise amplification level. Please refer to~\cite{ddnm} for method to derive $\kappa$ from noise level $\sigma$.}}
\end{equation}
\zjk{For nonlinear problems, we can directly solve the optimization objective in~\cref{eq:obj_piccs} by gradient descent (with momentum)~\cite{resample}. Finally, we simply skip this optional stage and rely on ControlNet to enforce measurement constraint if $\boldsymbol{A}$ is unknown or not differentiable.}

Multistep sampling serves as an iterative refinement process. 
In each step, we send the last-round sample back to a lower noise level by perturbing with a new noise. 
Then we denoise this newly perturbed sample with guided consistency model, resulting in an image with sharper and refined details.
By integrating hard measurement constraint with multistep sampling, we can obtain high-fidelity images more robustly.

\vspace{-6pt}
\section{Experiments}

\subsection{Experimental Settings}
\noindent\textbf{Tasks.} We evaluate our method across four distinct tasks, encompassing both linear and nonlinear inverse problems in the domains of natural and medical images. 
In natural image domain, we conduct experiments on two linear inverse problems: 4×Super-Resolution (SR) 
and block inpainting. 
Additionally, we also evaluate on nonlinear deblurring, illustrating the ability of our method to address nonlinear inverse problems. 
For medical image domain, we evaluate our method on sparse-view Computed Tomography (CT) reconstruction task, which aims at reconstructing CT images from under-sampled projections (sinograms). 
In our experiments, sinograms are simulated with Radon transformation using 23 projection angles equally distributed across 180 degrees. Following the baseline settings~\cite{score-sde, dps}, Gaussian noises with standard deviation $\sigma_{image}=0.05$ are added to the measurement space in experiments on natural images, while no additional noise is added on sinogram in CT reconstruction task.

\noindent\textbf{Datasets.} For natural image restoration tasks, we use the LSUN bedroom dataset~\cite{lsun}, which consists of over 3 million training images and 300 validation images. We leverage the consistency model checkpoint pre-trained on LSUN bedroom from ~\cite{cm}, and further train the ControlNet on the training set. For medical image restoration tasks, we utilize 2D slices sampled from AAPM LDCT dataset~\cite{ldct}. We train the model on a training set consisting of 3000 images from 40 patients, and test on 300 images from the remaining 10 patients. All images are in 256×256 resolution.

\noindent\textbf{Baselines.} In experiments on natural images, our baselines can be categorized into three main groups: (1)End-to-end image restoration methods like SwinIR~\cite{swinir} , (2)Plug-and-play diffusion-based inverse problem solvers (DIS), such as DDRM~\cite{ddrm} and DPS~\cite{dps}, and (3)Conditional diffusion-based method trained from scratch, such as I$^2$SB~\cite{i2sb} and CDDB~\cite{cddb}. We also compare with the zero-shot method based on unconditional consistency models proposed in ~\cite{cm} (denoted as CM). We omit CM and DDRM in nonlinear deblur since they require pseudo-inverse, which nonlinear operators do not have. In experiments on medical images, we compare our methods with: (1) Traditional mathematical methods like FBP, (2) Single-step deep reconstruction methods like FBP-UNet, and (3) Plug-and-play methods with diffusion or consistency model prior, such as MCG~\cite{mcg}, DPS~\cite{dps} and CM~\cite{cm}.

\noindent\textbf{Evaluation.} Considering varied demands in natural and medical image tasks, we employ different evaluation metrics for these two experiments.  For natural image tasks, we adopt perceptual metrics that align better with visual quality. Specifically,  we use Learned Perceptual Image Patch Similarity (LPIPS)~\cite{lpips} as a measurement of data consistency, and Fréchet Inception Distance (FID)~\cite{fid} as a measurement of image quality. \zjk{Since pixel-level metrics like Peak Signal-to-Noise Ratio (PSNR) and Structural Similarity Index (SSIM)~\cite{ssim} prefer blurry regression outputs~\cite{palette}, we leave those results in the Appendix.} Conversely, in medical image tasks where the risk of hallucination is undesirable, we use PSNR and SSIM as evaluation metrics instead of LPIPS and FID. These metrics align with our emphasis on similarity with the ground truth rather than focusing on image quality when processing medical images. 

\begin{table*}[t]
\centering
\begin{tabular}{llllllll}
\hline
                         &                        & \multicolumn{2}{c}{Block Inpainting} & \multicolumn{2}{c}{SR×4}        & \multicolumn{2}{c}{Nonlinear Deblur}                  \\ \cmidrule(l){3-4} \cmidrule(l){5-6} \cmidrule(l){7-8}
\multirow{-2}{*}{Method} & \multirow{-2}{*}{NFE↓} & LPIPS↓            & FID↓             & LPIPS↓         & FID↓           & LPIPS↓                & FID↓                          \\ \hline
SwinIR~\cite{swinir}                   & 1                      & 0.168             & 75.45            & \textbf{0.200} & 49.98          & 0.241                 & 84.12                         \\\hline
DDRM~\cite{ddrm}                     & 20                     & 0.223             & 51.50            & 0.283          & 54.39          & \multicolumn{1}{c}{-} & \multicolumn{1}{c}{-}         \\
CM~\cite{cm}                       & 39                     & 0.310             & 49.90            & 0.249          & 47.54          & \multicolumn{1}{c}{-} & \multicolumn{1}{c}{-}         \\
\rowcolor[HTML]{C0C0C0} 
DPS~\cite{dps}                      &   1000         & 0.245             & 43.54            & 0.226          & 46.15          & 0.291                 & 42.85                       \\\hline  
I$^2$SB~\cite{i2sb}                  & 2                      & 0.276             & 55.10            & 0.242          & 53.40          & 0.206                 & 48.40                         \\
\rowcolor[HTML]{C0C0C0} 
I$^2$SB~\cite{i2sb}                  & 999                    & 0.273             & 48.40            & 0.262          & 40.40          & 0.217                 & 44.50                         \\
CDDB~\cite{cddb}                     & 2                      & \textbf{0.126}    & 45.20            & 0.240          & 54.20          & 0.242                 & 64.90\\
\rowcolor[HTML]{C0C0C0} 
CDDB~\cite{cddb}                     & 999                    & 0.125& 40.50            & 0.236& 41.20& 0.231                 & 61.00                         \\ \hline
CoSIGN (ours)& 1                      & 0.146             & {\ul 39.89}      & 0.214    & {\ul 41.00}    & {\ul 0.185}           & {\ul 40.48}                   \\
CoSIGN (ours)& 2                      & {\ul 0.137}       & \textbf{38.64}   & {\ul 0.217}          & \textbf{40.84} & \textbf{0.167}        & \textbf{38.60}                \\ \hline
\end{tabular}
  \caption{Quantitative results of solving natural image inverse problems on LSUN bedroom validation set. Baselines using around 1000 NFEs are shadowed in grey and excluded for ranking. \textbf{Bold:} best; {\ul{underline}}: second best.}
  \vspace{-25pt}
  \label{tab:natural}
\end{table*}

\begin{figure*}[t]
  \centering
   \includegraphics[width=\linewidth]{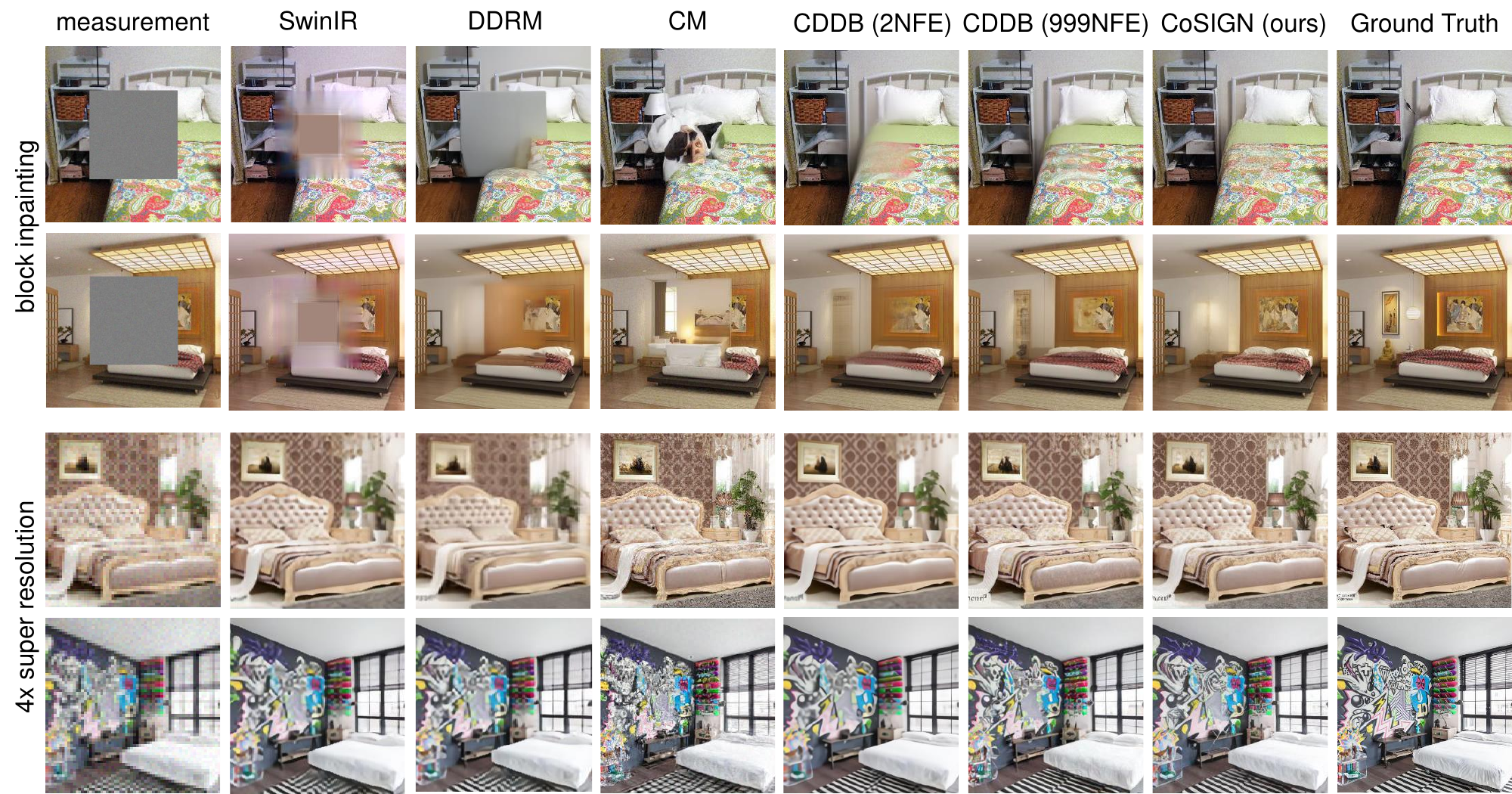}

   \caption{Visual results of two linear inverse problems on LSUN bedroom validation set. Zoom in to get a better view.}
   \vspace{-10pt}
   \label{fig:natural}
\end{figure*}

\subsection{Results on Natural Image Tasks}

\begin{figure}[t]
  \centering
   \includegraphics[width=\linewidth]{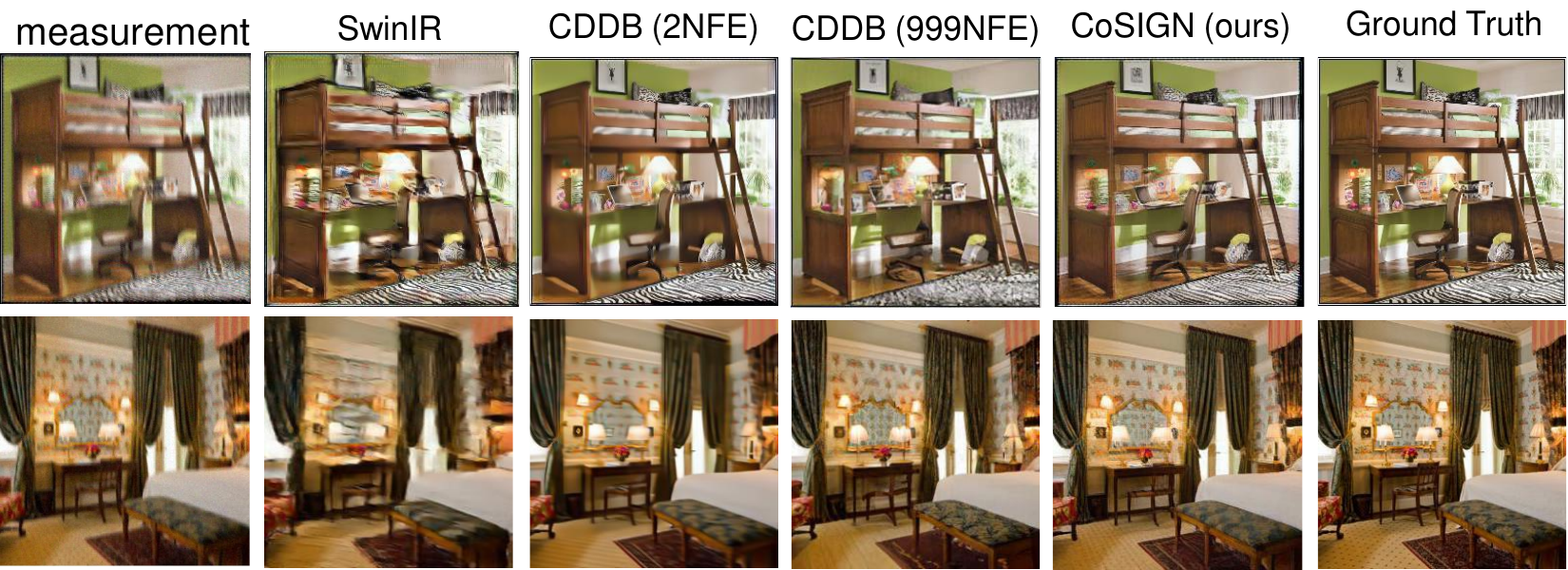}
   \vspace{-18pt}
   \caption{Visual results of nonlinear deblurring on LSUN bedroom validation set. Zoom in to get a better view.}
   \vspace{-20pt}
   \label{fig:nonlinear}
\end{figure}

We quantitatively and qualitatively compare our method and the baselines in \cref{tab:natural},~\cref{fig:natural} and~\cref{fig:nonlinear}, respectively. We first compare our method with baselines using NFE$<$100. Aside from methods working in high NFE region, our 2-step results surpass existing methods in most tasks with both metrics. 
Our 2-step and single-step results achieve the best and the second-best FID respectively across all tasks. This demonstrates the superior image quality of our reconstruction results. \zjk{It is worth noticing that CoSIGN works exceptionally well on nonlinear deblur, showcasing its ability to solve nonlinear inverse problems without pseudo-inverse.} Meanwhile, we would like to clarify that the sub-optimal performance of our method in LPIPS does not indicate data inconsistency. As depicted in~\cref{fig:natural}, although single-step results of SwinIR and 2-step results of CDDB suffer from obvious over-smoothness, they outperform our method in terms of LPIPS values. We suppose that this is because LPIPS hardly tolerate hallucination, which is essential for generating high-fidelity details. Additionally, we also compare our method with baselines resulted from high NFEs. As shadowed in grey in~\cref{tab:natural}, our method achieves performance on par with or even better than these baselines by using much less NFE steps. In conclusion, CoSIGN successfully "concentrate" the power of diffusion model prior into a single inference step or two, accelerating the process of inverse problem solving.



\subsection{Results on Medical Image Tasks}

\begin{figure}[t]
  \centering
   \includegraphics[width=\linewidth]{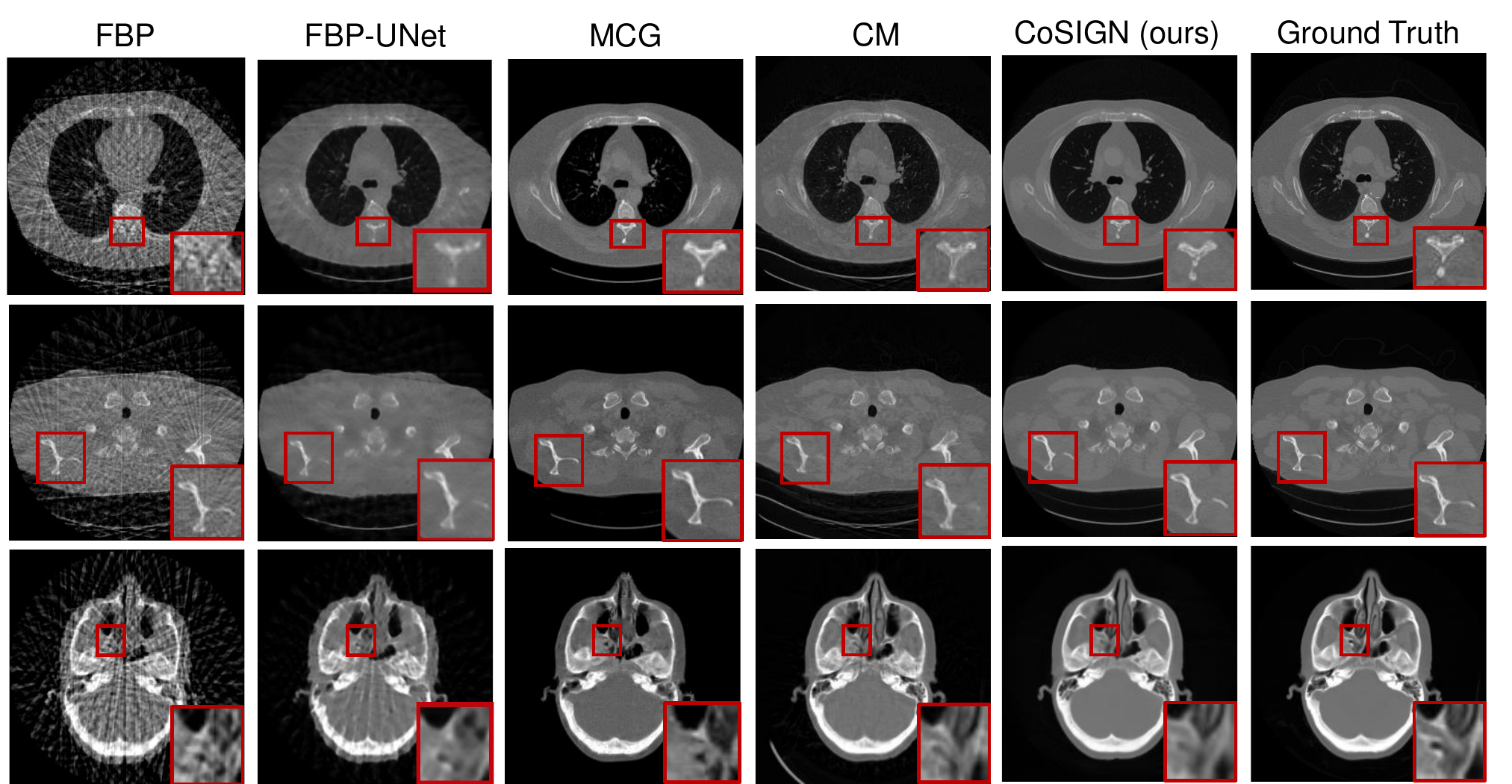}
   \vspace{-18pt}
   \caption{Visual results of sparse-view CT reconstruction with 23 angles on LDCT validation set. A zoomed-in patch with details is provided in the corner.}
   \vspace{-10pt}
   \label{fig:medical}
\end{figure}

\begin{table}[t]
  \centering
  \begin{subtable}{0.49\textwidth}
    \centering
    \begin{tabular}{llll}
    \toprule
    Method     & NFE↓ & PSNR↑ & SSIM↑ \\ \midrule
    FBP        & 0    &   25.02    &   0.685    \\
    FBP-UNet~\cite{fbp_unet} & 1    &  31.47&  \textbf{0.874}\\\hline
    CM~\cite{cm}& 39&       26.55&       0.737\\
    \rowcolor[HTML]{C0C0C0} MCG~\cite{mcg}        & 1000 &  29.48     &  0.847     \\
    \rowcolor[HTML]{C0C0C0} DPS~\cite{dps} & 1000    & 25.52   &   0.697    \\
    \hline
    CoSIGN (ours)       & 1    &       {\ul33.41}&       0.836\\
    CoSIGN (ours)       & 2    &       \textbf{34.26}&       {\ul0.866}\\ \bottomrule
    \end{tabular}
    \caption{}
    \label{tab:medical}
  \end{subtable}
  \hfill
  \begin{subtable}{0.49\textwidth}
    \centering
    \begin{tabular}{@{}lllll@{}}
    \toprule
    CM         & ControlNet & \begin{tabular}[c]{@{}l@{}}Guidance \\ Scale\end{tabular} & LPIPS↓ & FID↓ \\ \midrule
               & \checkmark & -             & 0.153  & 42.68\\
    \checkmark &            & 0              &        0.729&      43.34\\
    \checkmark & \checkmark & 0.3            &        0.421&      46.32\\
    \checkmark & \checkmark & 0.6            &        0.208&      40.46\\
    \checkmark & \checkmark & 1.0            & \textbf{0.146}  & \textbf{39.89}\\ \bottomrule
    \end{tabular}
    \caption{}
    \label{tab:ablation}
  \end{subtable}
  \vspace{-10pt}
  \caption{(a) Quantitative results of CT reconstruction with 23 angles on LDCT validation set. Baselines using 1000 NFEs are shadowed and excluded for ranking. \textbf{Bold:} best; {\ul{underline}}: second best; (b) Ablation study on consistency model prior and ControlNet. When guidance scale decrease to 0, the model will degenerate to unconditional consistency model. Setting guidance scale to 1.0 will recover our original method.}
  \label{tab:bothTables}
  \vspace{-30pt}
\end{table}

We report the quantitative and qualitative results on medical image tasks in \cref{tab:medical} and~\cref{fig:medical}, respectively. 
Our method surpasses all baselines with low NFEs in terms of PSNR metric. Whereas for SSIM, we fall behind FBP-UNet. 
But as showcased in~\cref{fig:medical}, our method generates sharper details while preserving high consistency with measurements compared to baseline methods such as FBP-UNet. 
For baselines with 1000 NEFs, our 2-step results are still comparable in the sense of both metrics and visual quality. Compared with baselines like MCG~\cite{mcg}, our method particularly excel in reconstructing clear soft tissues (\cref{fig:medical}).


\subsection{Ablation Study}

Ablation study focuses on three main components of our method: consistency model prior, ControlNet and hard measurement constraint. \zjk{Ablation on CM prior and ControlNet is conducted on block inpainting, whereas ablation on hard measurement constraint is conducted on CT reconstruction.} 

\begin{figure}[t]
  \centering
   \includegraphics[width=\linewidth]{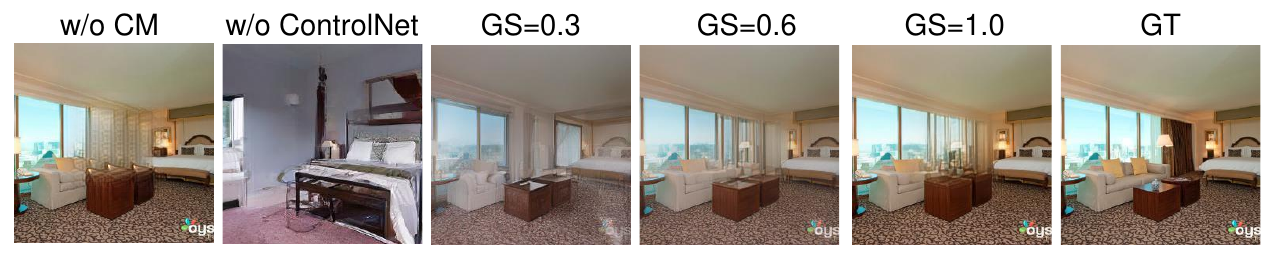}
    \vspace{-20pt}
   \caption{Visual quality of samples generated with different guidance scales, showing how condition is gradually injected with soft measurement constraint.}
   \vspace{-18pt}
   \label{fig:ablation}
\end{figure}

\noindent\textbf{Consistency Model Prior.} In order to corroborate the effectiveness of consistency model prior, we compare the single-step result with an end-to-end supervised model directly trained to transform pseudo-inverse of measurements into high-fidelity images. Like ControlNet, the end-to-end model has the same structure as U-Net in the consistency model and is trained with the same LPIPS loss. Its parameters are initialized with the same consistency model checkpoint as well. As shown in the first row of ~\cref{tab:ablation} and the first column in ~\cref{fig:ablation}, the end-to-end model without consistency model prior (noted as ``w/o CM")) 
generates blurry result with distinct artifacts, which is alleviated when incorporating consistency model as a prior.

\noindent\textbf{ControlNet.} We demonstrate how ControlNet turns an unconditional consistency model prior into a conditional model by evaluating under different guidance scales. Specifically, we multiply the guidance scale with outputs of the ControlNet to manually tune the conditioning strength. In~\cref{tab:ablation}, we observed that quantitative results steadily improve as the guidance scale increases. Meanwhile, visual results in~\cref{fig:ablation} gradually transit from unconditional samples to measurement-consistent reconstruction results. This gives us an intuitive interpretation of how \zjk{ControlNet} guides the unconditional consistency model towards the ground truth.

\noindent\textbf{Sampling Steps and Hard Measurement Constraint.} In~\cref{sec:multistep}, we introduced hard measurement constraint and multistep sampling as refinement steps. In~\cref{tab:ablation_hard} we observe that these refinement steps indeed improve performance. We do not report NFE$>$2 since no further improvement in performance is observed. This might be a problem inherited from the CM backbone~\cite{cm,ctm}. 

\begin{table}[htbp]
  \centering
  \begin{subtable}{0.3\textwidth}
    \centering
    \begin{tabular}{@{}llll@{}}
    \toprule
    NFE & HMC & PSNR↑          & SSIM↑           \\ \midrule
    1   &     & 31.38          & 0.8117          \\
    2   &     & 31.80          & 0.8298          \\
    1   & \checkmark  & 33.41          & 0.8355          \\
    2   & \checkmark  & \textbf{34.26} & \textbf{0.8660} \\ \bottomrule
    \end{tabular}
    \caption{}
    \label{tab:ablation_hard}
  \end{subtable}
  \hfill
    \begin{subtable}{0.6\textwidth}
    \centering
    \begin{tabular}{@{}llllll@{}}
    \toprule
    Method  & I$^2$SB   & CDDB   & Ours  & Ours   & Ours w/ HMC   \\ \midrule
    NFE     & 2      & 2      & 1     & 2      & 2      \\
    Time/ms & 111.24 & 121.42 & 69.53 & 139.11 & 139.47 \\ \bottomrule
    \end{tabular}
    \caption{}
    \label{tab:time}
  \end{subtable}
  \vspace{-10pt}
  \caption{(a) Ablation study on sampling steps and hard measurement constraint; (b) Per sample inference time. Measured on one single Nvidia A40 GPU. HMC is the abbreviation for hard measurement constraint.}
  \label{tab:bothTables_2}
  \vspace{-20pt}
\end{table}

\vspace{-6pt}
\section{Discussions}
\vspace{-3pt}

In this section, we will discuss two important issues \zjk{of CoSIGN}: the exact inference time, \zjk{as well as performance on out-of-domain (OOD) inverse tasks and noise scales}.

\noindent\textbf{Inference Time.} We report the per-image inference time of block inpainting on a single Nvidia A40 GPU in ~\cref{tab:time}. Our single-step reconstruction method can generate an image within 100ms, enabling real-time applications like video interpolation. Our two-step generation speed is similar to I$^2$SB~\cite{i2sb} and CDDB~\cite{cddb} and is also magnitudes faster than most previous diffusion-based methods. 

\zjk{\noindent\textbf{Out-of-domain (OOD) Adaptability.} To demonstrate the robustness of CoSIGN, we test our method on OOD inverse tasks and noise scales. The model is trained on sparse-view CT reconstruction with 23 angles, and no noise is added to the sinogram during training. Then we test the model's generalizability by conducting experiments on OOD tasks only using 10 angles for reconstruction (see~\cref{tab:ood_task}). Whereas in experiments on OOD noise scales, noises with different derivation $\sigma$ were added to the sinogram (see~\cref{fig:ood_noise}). As the number of projections decrease from 23 to 10, PSNR of our method drop slower than FBP-UNet. Meanwhile, performance of our method was much less affected by OOD noise. This indicate that the frozen CM prior endows our method with better robustness to OOD measurements compared with traditional approaches.}

\begin{table}[htbp]
\vspace{-10pt}
  \centering
  \begin{minipage}[b]{0.45\textwidth}
    \centering
    \begin{tabular}{@{}lllll@{}}
    \toprule
             & PSNR                        & SSIM                        \\ 
             \midrule
    FBP      & 17.23\scriptsize{-7.79}     & 0.488\scriptsize{-0.197}    \\ 
    FBP-UNet & 22.49\scriptsize{-8.98}     & 0.701\scriptsize{-0.173}    \\ 
    Ours     & \textbf{28.33}\scriptsize{-5.93} & \textbf{0.731}\scriptsize{-0.135} \\
    \bottomrule
    \end{tabular}
    \vspace{25pt}
    \caption{Robustness of our method to OOD inverse tasks. Decrease from in distribution results (23 angles) in \scriptsize{small}.}
    \label{tab:ood_task}
  \end{minipage}
  \hfill
  \begin{minipage}[b]{0.5\textwidth}
    \centering
    \includegraphics[width=\textwidth]{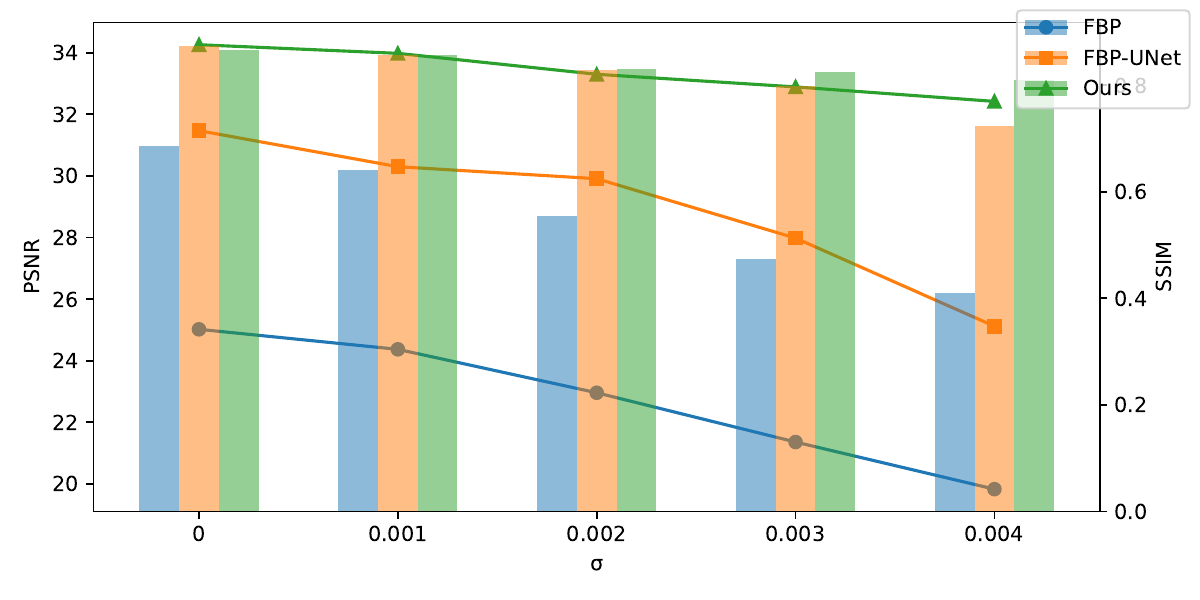}
    \caption{Robustness of our method to OOD noise scales. Polylines represent PSNR whereas bars represents SSIM.}
    \label{fig:ood_noise}
  \end{minipage}
  \vspace{-45pt}
\end{table}

\section{Conclusion}\label{sec:conclusion}
\vspace{-3pt}
In this work, we propose CoSIGN, a few-step inverse problem solver with consistency model prior. 
We propose to guide the conditional sampling process of consistency models with both soft measurement constraint and hard measurement constraint, which enables generating high-fidelity, measurement-consistent reconstructions within 1-2 NFEs.
Extensive experiments demonstrate our superiority against existing supervised and unsupervised diffusion-based methods under few-step setting.
\bibliographystyle{splncs04}
\bibliography{main}

\clearpage
\setcounter{page}{1}

%
%
%

\appendix
\setcounter{section}{0}
\renewcommand{\thesection}{\Alph{section}}

\renewcommand{\thesubsection}{\thesection.\arabic{subsection}}

\section{Limitations}
Compared to traditional diffusion-based inverse problem solvers (DIS), CoSIGN reduces number of sampling steps to 1-2 NFEs. However, the training of a ControlNet for each inverse task may limit the generalizability of the proposed method. Although we demonstrated its robustness against number of angles and noise scales in sparse-view CT reconstruction, a performance gap still exists when adapting the trained ControlNet to a different task. Future works may explore ways to utilize few-shot adaptation method for the training of ControlNet, or improve zero-shot inference ability of the proposed method.

\section{Implementation Details}

We implement our proposed algorithm (CoSIGN) based on the consistency model (CM) codebase\footnote{\url{https://github.com/openai/consistency_models}} so that we can make use of the CM checkpoint pretrained on the LSUN bedroom dataset~\cite{lsun}. The UNet structure of CM contains 6 resolution levels for the input size of 256×256. There are two residual blocks for each resolution level in both the encoder and the decoder. In the architecture of the additional encoder for guiding the CM backbone with the conditional input, we replaced each decoder layer with a zero-initialized convolution layer. We also add a zero-convolution layer before the additional encoder. We maintain the middle block in CM at the end of the additional encoder. The output of the middle block will pass through a zero-initialized convolution layer before entering the CM. We inject these conditions into CM by directly adding them with the skip connections between the encoder and the decoder. For medical images, we change the input channel of the first layer in both CM and the additional encoder into one.

In experiments on natural images, we train the additional encoder for 50k steps with a batch size of 144. In experiments on medical images, we start from training the diffusion model since no pretrained checkpoint is available. Specifically, we first train an EDM~\cite{edm} model on LDCT training set~\cite{ldct} for 9k steps with a batch size of 144. Then we distill this diffusion model into a consistency model by training for another 12k steps. Finally, we train the additional encoder for 9k steps with the CM backbone frozen. We do not train these models for further steps since it might induce over-fitting on such a small dataset.

We adopt the forward operator of different inverse problems from DPS codebase\footnote{\url{https://github.com/DPS2022/diffusion-posterior-sampling}} and add hard measurement constraints like DDNM~\cite{ddnm} into it. 

For evaluation, we adopt codes from DPS~\cite{dps} to calculate PSNR and SSIM whereas codes from CM~\cite{cm} to calculate FID. Following~\cite{cm}, the intermediate noise level is determined by ternary search in multistep sampling.


\begin{figure}[t]
  \centering
   \includegraphics[width=0.8\linewidth]{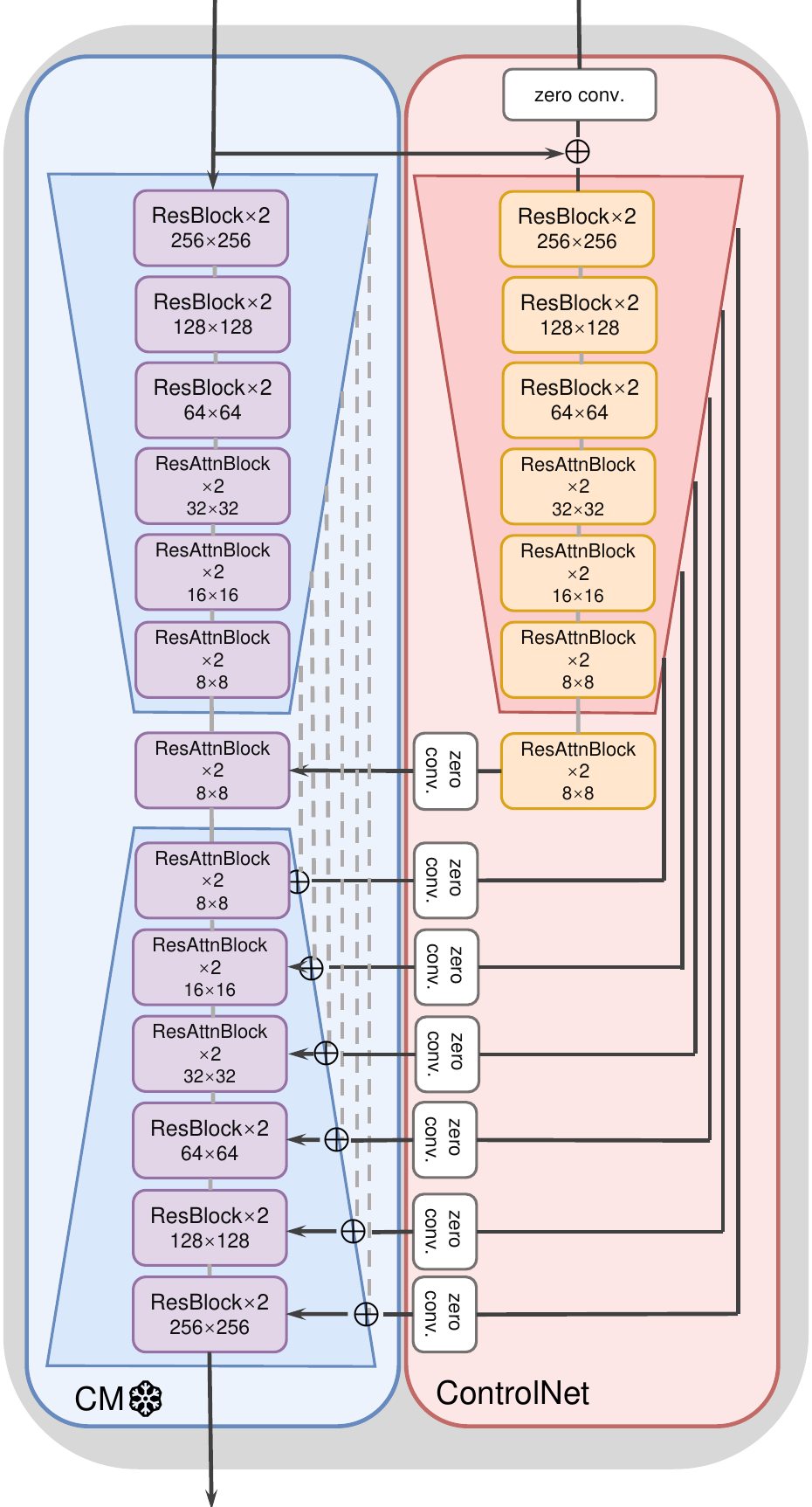}
   \caption{Illustration of our proposed CoSIGN model structure. ``ResAttnBlock×2" denotes a ``ResBlock-Attention Block-ResBlock" structure.}
   \vspace{-9pt}
   \label{fig:model_appendix}
\end{figure}

\section{Implementation Details of Baselines}
\noindent\textbf{SwinIR}

For super-resolution, we follow the default setting in 4x superresolution used in the original codebase provided by~\cite{swinir}, and trained for 500k iterations on the LSUN-bedroom training set. For nonlinear deblurring and box-inpainting, we train swinir by mapping the degraded images to the ground truth images for 500k iterations. 


\paragraph{DPS and MCG.} For DPS and MCG, we use the original codebase provided by~\cite{dps, mcg} and pre-trained DDPM models~\cite{ddpm} trained on LSUN-bedroom training sets. We follow the default setting with NFE being 1000.

\paragraph{DDRM.} For DDRM, we follow the original code provided by \cite{ddrm} with DDPM models trained on LSUN-bedroom training sets. We use the default parameters as displayed by \cite{ddrm} with NFE being 20.

\noindent\textbf{CM} As~\cite{cm} did not report quantitative results on inverse problem solving tasks, we used the iterative inpainting and the iterative super-resolution functions in their codebase to reproduce their results. We try to keep measurement consistency and improve image quality by maximizing the number of iterations.

\noindent\textbf{I$^2$SB, CDDB} The original I$^2$SB model was trained on ImageNet. To compare it with our method, we fine-tuned it on LSUN-bedroom for 6k steps with a batch size of 256. We initialized the model for nonlinear deblur task using the checkpoint for Gaussian deblur task since no checkpoint is available on this task. Experiments on CDDB is also re-conducted on these fine-tuned models.

\noindent\textbf{FBP-UNet}
For FBP-UNet, we use the model structure as described in~\cite{fbp_unet} and then train the model with input images being FBP reconstructions and output being ground truth images of the 9000 2D CT slices from 40 patients. 

\section{Additional Results}
\subsection{Distortion Metrics}
In~\cref{tab:psnr}, we present distortion metrics of three natural image restoration tasks. It should be noticed that unlike CT reconstruction with 23 angles, these inverse tasks are considerably aggressive. To make up for information loss while maintaining image quality, some degree of hallucination is necessary~\cite{palette,dps}. However, PSNR/SSIM strictly penalize hallucination as they rely on pixel-level differences.

We would like to clarify that DDB methods~\cite{i2sb,cddb} outperform ours in low NFE region not because they produce higher-fidelity images, but because they "trade accuracy with quality". When working within 1-2 NFEs, DDB methods generate samples closer to $\mathbb{E}[\boldsymbol{x}_0|\boldsymbol{x}_t]$ rather than any clear $\boldsymbol{x}_0$ from the real distribution $p(\boldsymbol{x})$. As shown in~\cref{fig:natural} and~\cref{fig:nonlinear}, methods with superior distortion metrics mostly generate blurry samples, indicating that they seek the mean of all possible reconstructions rather than a single clear result.

Having acknowledged this, we agree that distortion may be detrimental in certain inverse tasks, especially those with medical applications. Therefore, we provide distortion metrics of medical image reconstruction tasks in~\cref{tab:medical}, and those of natural image reconstruction tasks in~\cref{tab:psnr} for reference.

\begin{table}[]
\centering
\begin{tabular}{llllllll}
\hline
        &      & \multicolumn{2}{l}{Block Inpainting}                              & \multicolumn{2}{c}{SR×4}                                       & \multicolumn{2}{c}{Nonlinear Deblur}                              \\
        &      & PSNR↑                           & SSIM↑                           & PSNR↑                        & SSIM↑                           & PSNR↑                           & SSIM↑                           \\
SwinIR  & 1    & 20.21                           & 0.796                           & 27.04                        & 0.805                           & 23.32                           & 0.685                           \\ \hline
DDRM    & 20   & 18.90                           & 0.624                           & 24.95                        & 0.691                           & \multicolumn{1}{c}{-}           & \multicolumn{1}{c}{-}           \\
CM      & 39   & 18.16                           & 0.660                           & 24.91                        & 0.742                           & \multicolumn{1}{c}{-}           & \multicolumn{1}{c}{-}           \\
DPS     & 1000 & 18.93                           & 0.630                           & 25.07                        & 0.723                           & 24.68                           & 0.702                           \\ \hline
I$^2$SB & 2    & {\ul 23.21}    & 0.715                           & {\ul 27.23} & {\ul 0.816}    & {\ul 28.30}    & {\ul 0.843}    \\
\rowcolor[HTML]{C0C0C0} 
I$^2$SB & 999  & 20.68                           & 0.685                           & 24.63                        & 0.721                           & 26.78                           & 0.792                           \\
CDDB    & 2    & \textbf{23.74} & \textbf{0.859} & \textbf{27.31} & \textbf{0.819} & \textbf{28.51} & \textbf{0.847} \\
\rowcolor[HTML]{C0C0C0} 
CDDB    & 999  & 22.97                           & 0.847                           & 25.27                        & 0.740                           & 27.80                           & 0.836                           \\ \hline
Ours    & 1    & 22.28                           & 0.828                           & 25.38                        & 0.764                           & 25.75                           & 0.791                           \\
Ours    & 2    & 22.61                           & {\ul 0.841}    & 26.13                        & 0.769                           & 26.86                           & 0.816                           \\ \hline
\end{tabular}
\label{tab:psnr}
\caption{Distortion metrics of solving natural image inverse problems on LSUN bedroom validation set. Baselines using around 1000 NFEs are shadowed in grey and excluded for ranking. \textbf{Bold}: best; {\ul underline}: second best. }
\end{table}

\subsection{Results on Natural Image Restoration}

\begin{figure*}[t]
  \centering
   \includegraphics[width=0.9\linewidth]{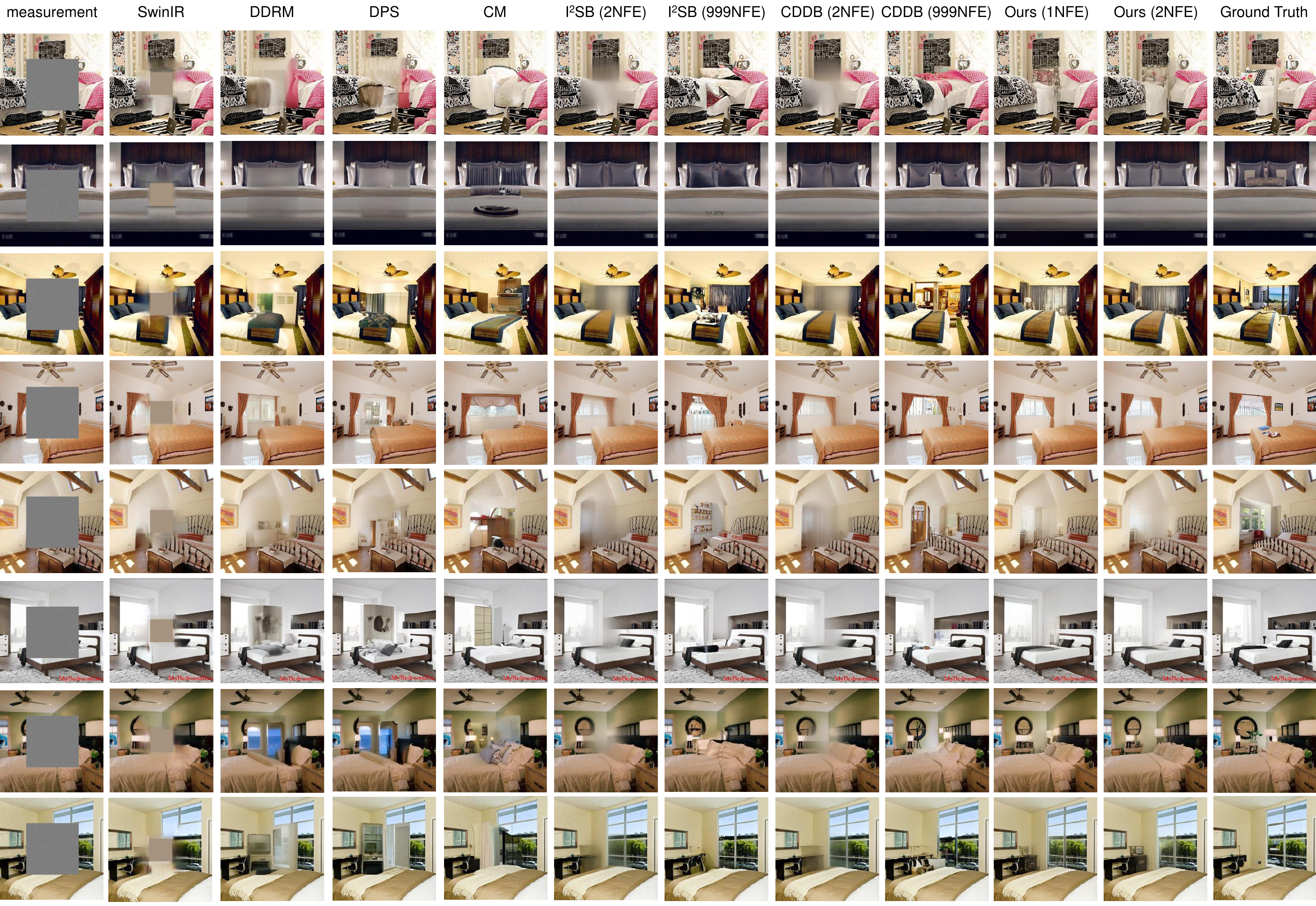}
   \vspace{-10pt}
   \caption{Additional results on central block inpainting on LSUN bedroom validation set.}
   \vspace{-10pt}
   \label{fig:inpaint}
\end{figure*}

\begin{figure*}[t]
  \centering
   \includegraphics[width=0.9\linewidth]{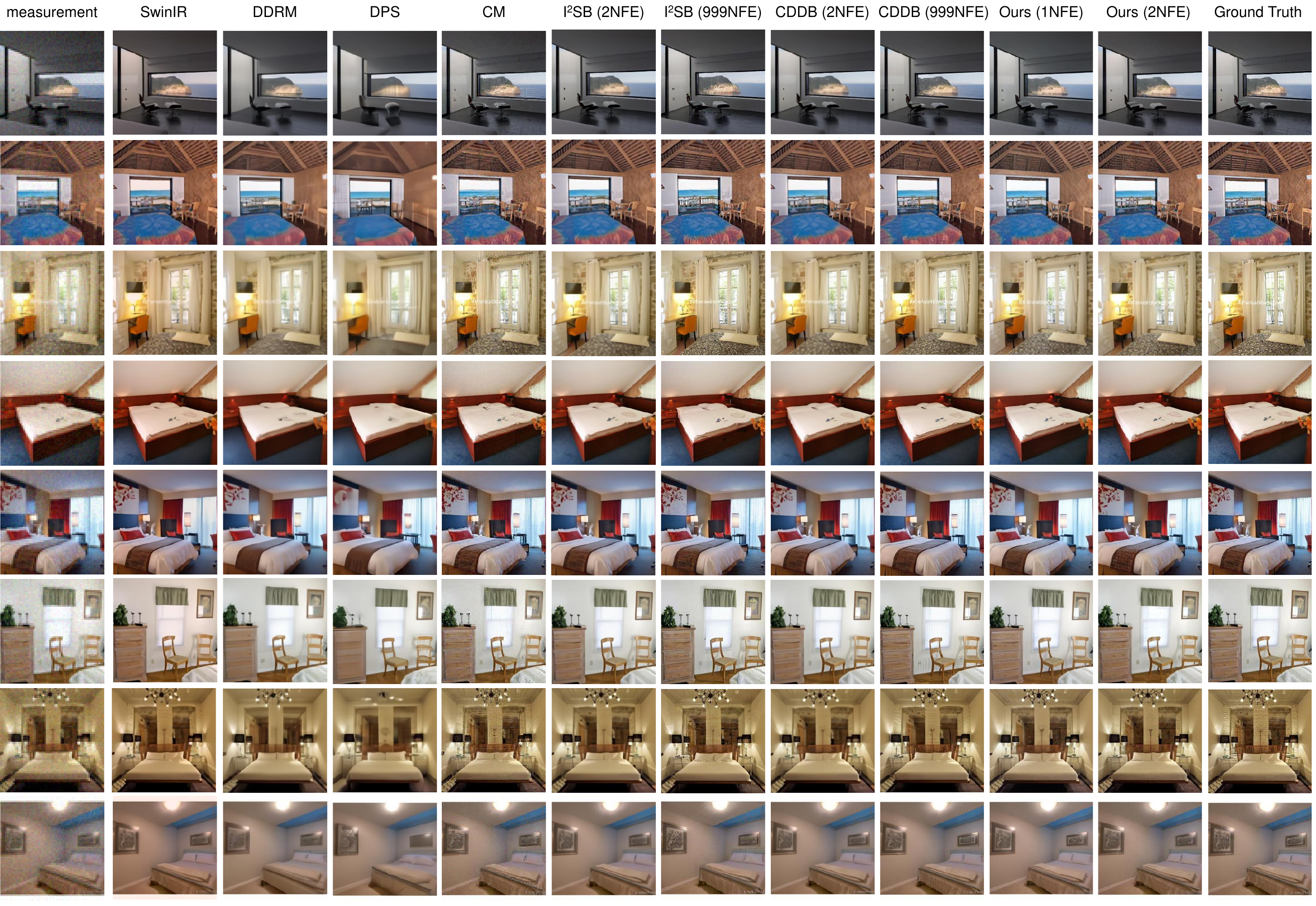}
   \vspace{-10pt}
   \caption{Additional results on super-resolution on LSUN bedroom validation set.}
   \vspace{-10pt}
   \label{fig:sr}
\end{figure*}

\begin{figure*}[t]
  \centering
   \includegraphics[width=\linewidth]{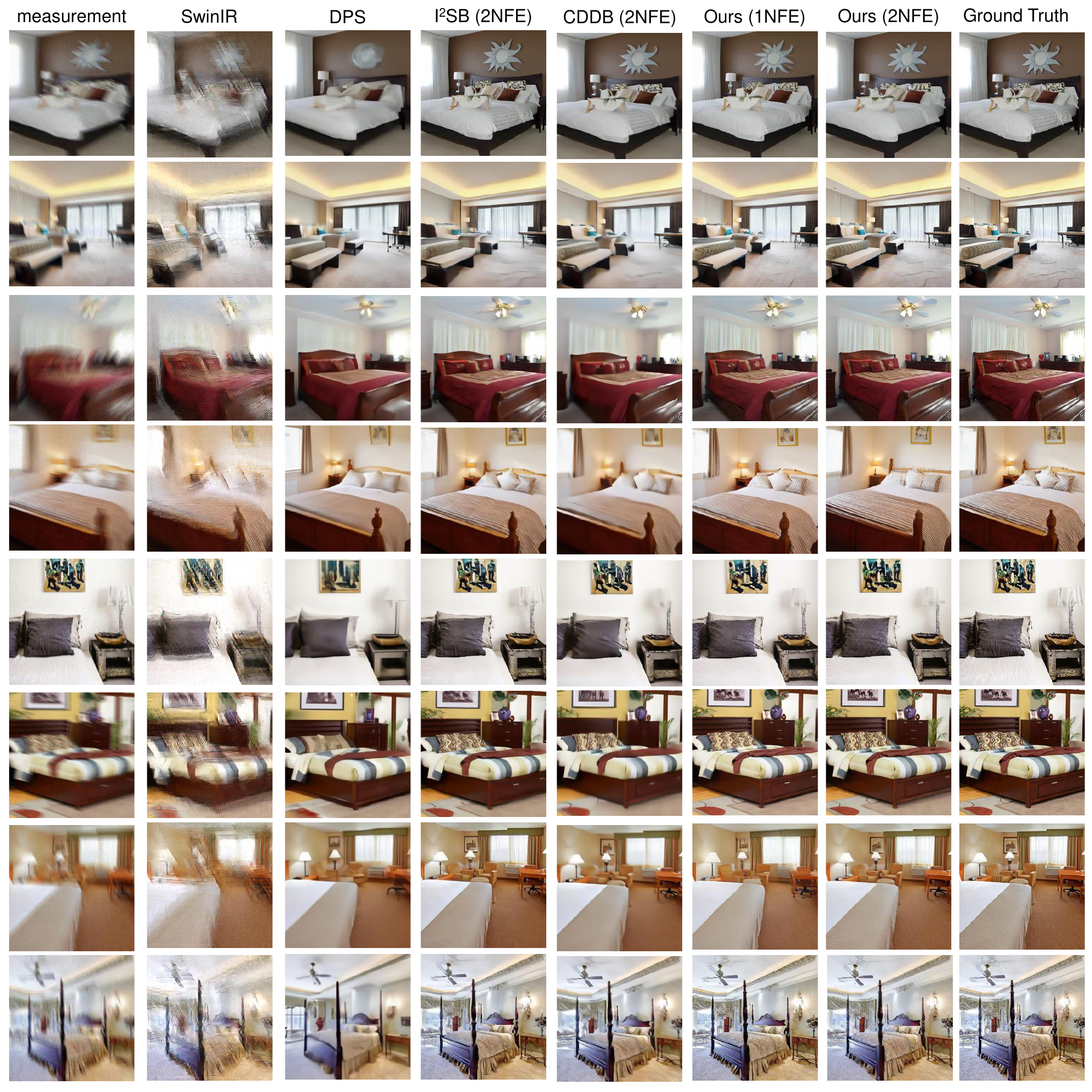}
   \caption{Additional results on nonlinear-deblur on LSUN bedroom validation set.}
   \label{fig:nl}
\end{figure*}

We provide additional visual results on natural image restoration of both CoSIGN and the baselines in ~\cref{fig:inpaint}, ~\cref{fig:sr} and ~\cref{fig:nl}. All images are randomly selected from the dataset without cherry picking. As depicted in these images, the visual quality of our results surpasses all existing methods in comparable NFE region, and is also comparable with those obtained with hundreds of NFEs.

\subsection{Results on Medical Image Restoration}

\begin{figure*}[t]
  \centering
   \includegraphics[width=\linewidth]{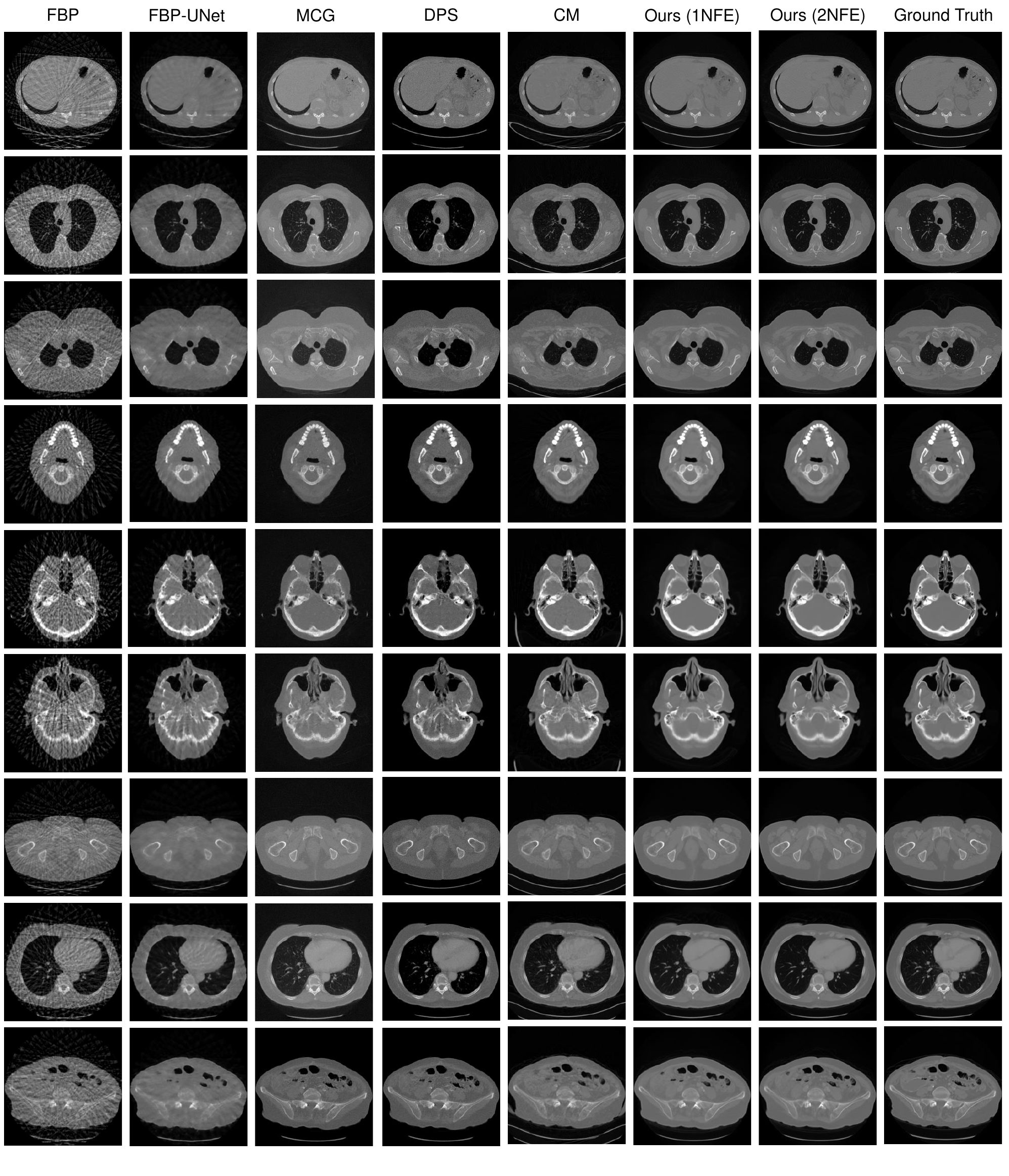}
   \vspace{-18pt}
   \caption{Additional results on sparse-view CT reconstruction with 23 angles on LDCT validation set.}
   \vspace{-18pt}
   \label{fig:medical_appendix}
\end{figure*}

In ~\cref{fig:medical_appendix}, we provide additional visual results on medical image restoration of both CoSIGN and the baselines. The selected images encompass CT scans of abdomen, head and chest. It can be seen from the images that compared with baselines, images reconstructed with CoSIGN are both high-fidelity and noiseless.

\subsection{Derivation of Hard Consistency Formula in the Linear and Noiseless Case}
If the forward operator $\mathbf{A}$ is linear, full-rank and the measurements are noiseless (i.e., $\boldsymbol{y} = \mathbf{A}(\boldsymbol{x})$), suppose we want to find the closest point to $\boldsymbol{x}_0$ that is consistent to the measurement $\boldsymbol{y}$, then we can pose the optimization problem as

\begin{equation}
\widehat{\boldsymbol{x}}_{0}=\underset{\boldsymbol{z} \in \mathbb{R}^n}{\arg \min}\left\{\|\boldsymbol{z}-\boldsymbol{x}_0\|_2^2\right\} \quad \text{s.t. } \mathbf{A}(\boldsymbol{z}) = \boldsymbol{y}
  \label{eq:obj_piccs1}
\end{equation}

Then, the solution to this optimization problem is given by 
\begin{align}
\hat{\boldsymbol{x}}_0 = \boldsymbol{x}_0 - (\mathbf{A}^{+}\mathbf{A}\boldsymbol{x}_0 - \mathbf{A}^{+}\boldsymbol{y}),
\end{align}
\textbf{Proof:}
Consider $\mathbf{t} = \boldsymbol{z} - \mathbf{x}_0$, then the previous optimization objective can be written to 
\begin{equation}
\widehat{\boldsymbol{x}}_{0}=\underset{\boldsymbol{z} \in \mathbb{R}^n}{\arg \min}\left\{\|\boldsymbol{t}\|_2^2\right\} \quad \text{s.t. } \mathbf{A}(\boldsymbol{t}) = \boldsymbol{y} - \mathbf{A}\mathbf{x_0}
  \label{eq:obj_piccs2}
\end{equation}
Then we can decompose $\mathbf{t}$ into a null space component and a perpendicular range space component, such that $\mathbf{t} = \mathbf{t}_{N(\mathbf{A})} + \mathbf{t}_{R(\mathbf{A}^T)}$, where $N(\mathbf{A}) \perp R(\mathbf{A}^T)$. We also have $\mathbf{A}\boldsymbol{t} = \mathbf{A}\mathbf{t}_{R(\mathbf{A}^T)} = \mathbf{A}\mathbf{A}^Tk =\boldsymbol{y} - \mathbf{A}\mathbf{x_0}$, by $\mathbf{t}_{R(\mathbf{A}^T)} = \mathbf{A}^Tk$. Then $k = (\mathbf{A}\mathbf{A}^T)^{-1}(\boldsymbol{y} - \mathbf{A}\mathbf{x_0})$, and then $\mathbf{t}_{R(\mathbf{A}^T)} = \mathbf{A}^T(\mathbf{A}\mathbf{A}^T)^{-1}(\boldsymbol{y} - \mathbf{A}\mathbf{x_0}) = \mathbf{A}^\dag (\boldsymbol{y} - \mathbf{A}\mathbf{x_0})$

We also have $||\boldsymbol{t}||_2^2 = ||\mathbf{t}_{N(\mathbf{A})}||_2^2 + ||\mathbf{t}_{R(\mathbf{A}^T)}||_2^2$, observe when $\mathbf{t}_{N(\mathbf{A})} = 0$,$||\boldsymbol{t}||_2^2$ is minimized. Hence,  $\boldsymbol{z} = \boldsymbol{x}_0 + \mathbf{A}^\dag (\boldsymbol{y} - \mathbf{A}\mathbf{x_0})$. 

\end{document}